%% file: main.tex
\documentclass{article} 
\usepackage{iclr2026_conference,times}

\input{math_commands.tex}

\usepackage{hyperref}
\usepackage{url}
\usepackage{booktabs}       
\usepackage{amsfonts}       
\usepackage{nicefrac}       
\usepackage{microtype}      
\usepackage{xcolor}      
\usepackage{todonotes}
\usepackage{amsmath}
\usepackage{graphicx}
\usepackage{amssymb}
\usepackage{mathtools}
\usepackage{amsthm}
\usepackage{subcaption}
\usepackage{siunitx}

\newcounter{spcounter}\renewcommand{\thespcounter}{\arabic{spcounter}}
\newcommand{\speciallabel}[1]{\refstepcounter{spcounter}\label{#1}(\thespcounter)}
\DeclareSIUnit\watthour{\watt\hour}

\title{Spiking Neural Networks for Continuous Control: Neuromorphic Reinforcement Learning in Conventional Computing}

\author{Jessica Hunter\\
Department of Computer Science\\
New Mexico Institute of Mining and Technology\\
Socorro, NM 87801, USA \\
\texttt{jessica.hunter@student.nmt.edu} \\
\And
Md Maruf Hossain Shuvo \\
Department of Electrical and Computer Engineering \\
The University of Texas at El Paso \\
El Paso, TX 79968, USA \\
\texttt{mhshuvo@utep.edu} \\
\And
Krishna Roy \\
Department of Electrical Engineering \\
New Mexico Institute of Mining and Technology \\
Socorro, NM 87801, USA \\
\texttt{krisna.roy@nmt.edu} \\
}

\iclrfinalcopy 
\begin{document}

\maketitle

\begin{abstract}
 Reinforcement learning (RL) algorithms have made strides over the past decade applying them to a wide range of problems and control tasks. However, the deployment of RL on neuromorphic hardware for continuous control tasks remains under-validated. Namely it is unclear whether replacing a conventional actor network with a spiking neural network (SNN) affects the performance of an agent before any hardware-specific benefits manifest. We provide a systematic validation of a minimal, neuromorphically viable spiking actor variant of Soft Actor-Critic (SAC) on conventional hardware, establishing a baseline for future neuromorphic RL research. In this paper, we propose the Spiking Actor Network Soft Actor Critic (SANSAC) to address the use of RL frameworks in continuous environments, designed as a framework that can be implemented on neuromorphic hardware. We compare a traditional Soft Actor Critic (SAC) network to SANSAC in a traditional computer. We demonstrate the near equivalent performance of SANSAC and SAC, while addressing the impact of hidden dimensions. Our results demonstrate the viability of SNN based algorithms in complex continuous environments, as well as competitive performance to traditional neural networks in traditional computers, providing a basis to continue exploring the use of SNNs in continuous RL frameworks.
 \footnote{Implementation code is publicly available at \url{https://github.com/jhunter533/SANSAC}.}
\end{abstract}

\section{Introduction}
\label{Introduction}

Reinforcement learning (RL) as a field has drawn attention in recent years as breakthroughs continue \cite{10.5555/3540261.3540988, DBLP:journals/corr/abs-1912-01603}. Despite recent advancements made to the field, solutions for continuous control problems have stalled in respect to off-policy algorithms. The use of neuromorphic hardware including memristors makes this more evident. Neuromorphic hardware poses a problem for the implementation of traditional RL frameworks in continuous control problems. As continuous environments exacerbate the issue of reproducibility, stability, and portability found in RL frameworks \cite{pr10112311}. Progress has been slow due to the high complexity of tasks such as those seen in robotics consisting of Partially Observable Markov Decision Processes (POMDP). These tasks lead to an increase in resources needed to train and run an agent for a given task which creates a larger barrier to entry for solutions. Neuromorphic hardware attempts to resolve parts of these issues by decreasing energy cost, training time, and total hardware required for in-situ training on systems that traditionally use embedded hardware \cite{DBLP:journals/corr/abs-2003-01157}. This is a direct result of the truly parallel nature of these chips in comparison to traditional traditional computer architectures. However, current research is non extensive for solutions that are high performing on continuous control tasks implemented on neuromorphic hardware. Improvements in this area have been made as the prioritization of spiking neural networks (SNNs) increases, taking cues from biological process to steer cutting edge techniques which advance the ability of problem solving for continuous control problems.

Spiking neural networks, considered the third generation of artificial intelligence \cite{Maass1996NetworksOS}, have made large contributions to the implementation of neural networks on neuromorphic hardware. Differing degrees of biological plausibility have given rise to several neuron model types used by SNNs including Leaky-Integrate-Fire (LIF) \cite{8184808}, Hodgkin-Huxley \cite{https://doi.org/10.1113/jphysiol.1952.sp004764}, Izhikevich \cite{12683326}, and many more. In addition there exist several learning methods for networks including three-factor learning rules \cite{10.3389/fncir.2015.00085}, derived from biological cues, and surrogate gradients \cite{DBLP:journals/corr/abs-1901-09948}, derived from traditional gradient descent. The variety in techniques have allowed for expansion of the use of SNNs in neural network applications. With the acquisition of neuromorphic hardware proving difficult for most at present, it stands to reason that there is a clear gap in comparison of SNN and traditional neural networks on traditional computers. Understandably, this is a result of SNN benefits only being significant in neuromorphic hardware, however, there remains little in terms of RL to broadly assume this to be true for all cases. While it can reasonably assumed that SNNs would not perform significantly better than digital neural networks (DNNs) in this system, it remains an important confirmation to fully understand for continuous tasks. Through the observation of data comparisons on a traditional computer we can see the small differences SNNs and DNNs have that can cascade on individually advantaged hardware.

Implementation of neural networks on neuromorphic hardware, be it memristors \cite{Wang2019-oj} or the Intel Loihi \cite{8259423} require all network components to have a physical representation. Thus, models and designs must be constrained to reasonably available neurons. As this constraint poses a crucial block to rapid progress, we extend our research to compare DNN and SNN implementation in the same traditional computer architecture including observations of reduced neuron counts for all hidden layers to function operably.

We propose the implementation of the Spiking Actor Network Soft Actor Critic (SANSAC), a network designed to be a minimal modifcation of SAC where only the actor is replaced by an SNN. Our contribution is not the novelty of this suggestion but the systematic comparison of SANSAC to SAC on identical conventional hardware across multiple network sizes. We analyze the effect of hidden dimension on policy learning and quantify performance differences. This establishes a reproducible baseline against which future neuromorphic implementations can be compared. We aim to understand the core of SNNs in continuous control tasks by observing possible differences in several performance metrics across both algorithms.

\section{Related Work}
\label{Related Work}
Important to understanding the basis of our methods is the basics of neuromorphic hardware and alternative techniques that influenced the work.
\subsection{Population Encoding}
Present work in SNNs introduce the biologically plausible mechanism of population encoding \cite{article}. Described from neuroscience developments of the human brain's mechanisms, it was then applied to reinforcement learning frameworks \cite{10.1085/jgp.59.6.734}. While implementations differ from context to context \cite{SUN2025106898}, in continuous control environments it was popularized by PopSAN \cite{DBLP:journals/corr/abs-2010-09635, DBLP:journals/corr/abs-2003-01157} which create a general framework for the addition of population encoding and decoding to any traditional Actor-Critic algorithm. While this method for encoding and decoding closely mimics observed behavior of learning in the brain and improves associated SNN performance, we opt to not include it in our algorithm implementations. This stems from the added complexity in design and implementation as well as the possibility of it obscuring the data in our base algorithm comparison results.
\subsection{Neuromorphic Hardware}
Neuromorphic hardware encapsulates systems that aim to emulate biological processes principles through special hardware which is entirely separate from a classical traditional computer architecture. These systems employ a combination of analog, digital, and mixed-signal circuits to implement neurons and synapses. Unlike the traditional computer, these systems are driven by spike events. There has been growing interest in a areas that require limited resources \cite{9822407, 10.1063/1.5124027} to use this hardware. Whereas traditional computers are defined by their separate CPU and memory units creating bottlenecks for data transmission, neuromorphic machines use neurons and synapses allowing mass parallelization. Often done through the implementation of memristors or semiconductors \cite{516353733}. It is currently of growing interest to implement complex algorithms such as RL frameworks on these systems \cite{DBLP:journals/corr/abs-2001-06930} primarily in an effort to gain nanoscale size and lower power consumption \cite{9458542, Wang2019-oj}. The use of such hardware in complex problems including robotics and other embedded hardware systems can increase power and space usage while maintaining similar performance. Highly desired for future implementation of highly efficient sensory and decision making processes \cite{doi:10.1126/scirobotics.abl8419, YANG2023106838, 10473282}. The logical requirement of the system is that they rely on SNNs to cater to biologically plausible characteristics over traditional neurons in DNNs. To understand the impact of the design patterns associated with the hardware we use it to influence our algorithm design.

\section{Methods}
\label{Methods}

\subsection{Notation}
At the core of policy learning environments is the Markov decision processes (MDP) which defines the problem. We define the tuple derived from the MDP for our continuous action space environments as follows: $(s_{t},a_{t},r_{t},s_{t+1})$.  Where $s$ represents the state space at time step, $t$. $t+1$ is defined as the immediate following time step. We use $r$ to represent the reward of the state transition from the given action, $a$. We follow the standard notation of $\pi$ representing the policy with respect to action and state, $\pi (a|s)$. 
\begin{figure*}[h]
    \centering
    \includegraphics[width=0.95\linewidth]{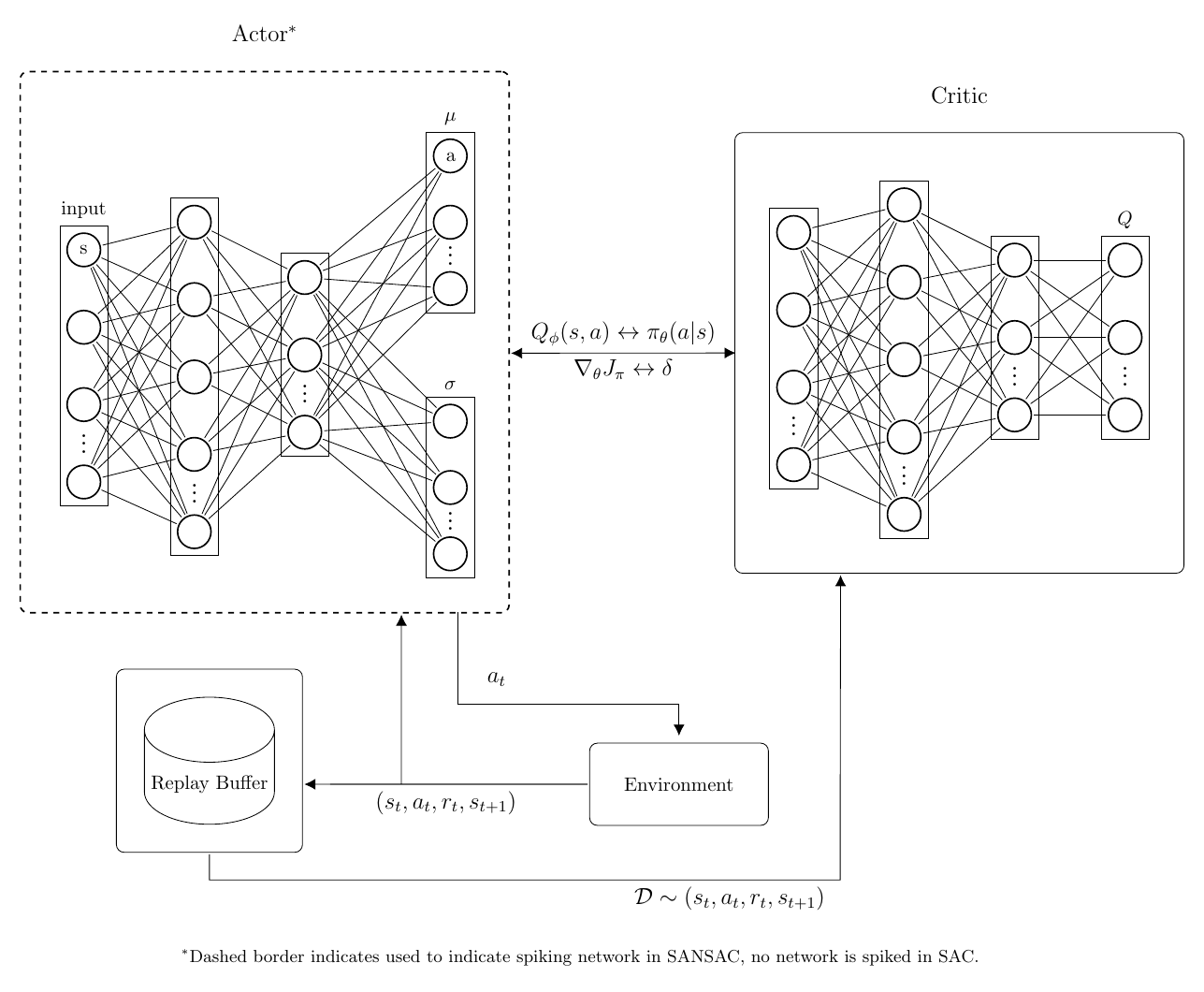}
    \caption{SANSAC/SAC Algorithm Diagram. Dashed border indicates spiking actor in SANSAC. All other components are identical between both algorithms. Critic outputs a state-action value $Q(s,a)$. Actor outputs a policy $\pi(a|s)$.}
    \label{fig:AlgDiagram}
\end{figure*}
\subsection{SAC}
The Soft Actor Critic (SAC) \cite{DBLP:journals/corr/abs-1801-01290}, is designed using stochastic policy and the maximum entropy objective, making it ideal for complex continuous environments. In addition, it has demonstrated high performance in the Farama Foundation Bipedal Walker environment \cite{towers2024gymnasium}. Environments such as Bipedal Walker are often prone to higher instability due to the complex continuous nature of them. This allows our strategy to be tested in a well established framework with similar problems to that of real world applications. In this work, we use "stability" to refer to the consistency of policy improvement across training episodes and the avoidance of training collapse such as catastrophic forgetting.

Instability of policy learning is in part influenced by the underlying equations of the algorithm. Key to SAC is the transformation required in the actor to acquire action samples, requiring logarithms to aid stochastic exploration. We derive the following more stable form of this transformation from Equation 21 in \cite{DBLP:journals/corr/abs-1801-01290}.

Starting from the original policy log-probability:
\begin{align}
    \log \pi(a|s) &= \log \mu(u|s) - \sum_{i=1}^{D} \log(1-\tanh^2(u_i)) \nonumber \\
    &= \log \mu(u|s) - 2\sum_{i=1}^{D} \log(\text{sech}(u_i)) \label{eq:policyStable}
\end{align}
where $\text{sech}(x)=2/(e^x+e^{-x})$ and $\mu(u|s)$ is the policy density over the pre-tanh variable $u$. This leverages the $\text{softplus}$ function for numerical stability during backpropagation. The equation derived reduces learning instability in complex continuous environments that have an increased susceptibility to numerical overflow, particular from $\tanh$ nonlinearity and Gaussian sampling operations. The full derivation of \autoref{eq:policyStable} appears in Appendix \ref{app:derivation}.
Implementation of SAC follows the configuration shown in \autoref{fig:AlgDiagram}, following closely to the presented logic of the original paper, where the algorithm consists of one actor network and two critic networks. Each network consists of three fully connected layers.
\subsection{SANSAC}
As SNNs have the capacity for physical neuromorphic hardware implementation, an algorithm inspired by the same framework necessitates implementation of spikes. Neuromorphic hardware has the capacity for advantages through the parallelization of data and computations. Through crossbar matrix multiplication simultaneous parallel calculations can be performed which can allow for lower energy costs and increased computation throughput. 

Spiking Actor Networks (SAN), inspired by the Actor-Critic framework, utilize an equivalent architecture to the deep learning model used as the basis with the modification that the actor network be made a SNN while the critics remain deep critic networks. Originally proposed in a series of papers by Tang et al, \cite{DBLP:journals/corr/abs-2003-01157,DBLP:journals/corr/abs-2010-09635}, this methodology became the basis of our algorithm configuration. This hybrid RL framework relies on the final deployable model consisting of the spiking network and discarding other components. The concept then allows for full deployment of a neural network on neuromorphic hardware without the need to overhaul the training process for complex algorithms such as SAC which depends on multiple critics during training time. Thus, the computational burden on a neuromorphic chip can be reduced and allow for easier scaling of networks.

Through the combination of the SAC and SAN frameworks we propose the Spiking Actor Network Soft Actor Critic (SANSAC). The deep neural network is maintained for the two critics while the actor network is implemented as a spiking neural network. The replay buffer equally critical to this framework remains the same as in a traditional SAC and is discarded after the completion of the training phase. In theory, this implementation would allow for a hybrid training system and fully neuromorphic agent after the training.

Spiked neural networks do not rely upon activation functions but rather rely on neurons with dynamics described by complex equations. As such SANSAC relies on the widely chosen LIF neuron. The LIF model is characterized by the following differential equation to describing the temporal evolution of the membrane potential:
\begin{equation}
    \tau_{\mathrm{m}} \frac{\mathrm{d}V}{\mathrm{d}t} = -(V - V_{\mathrm{rest}}) + RI
    \label{eq:LIF}
\end{equation}
where $\tau_{\mathrm{m}}$ is the membrane time constant, $V_{\mathrm{rest}}$ is the resting potential, $R$ is the membrane resistance, and $I$ is the input current.

Traditional output of the LIF neuron layer is a spike train. This provides a hindrance to the decoding of network outputs into a continuous action space. An alternative proposed \cite{chen2024fullyspikingactornetwork},\cite{DBLP:journals/corr/abs-2201-09754} relies on modifying the output neuron layer form LIF to a nonspiking LIF layer referred to as Leaky-Integrate (LI) neurons. These output a constant voltage over the traditional spike train, which is key for decoding in continuous action spaces. SANSAC makes use of this through the final layer in the actor network which uses a LI neuron configuration while the rest of the layers use LIF neurons. 

The implementation of SANSAC is accomplished through the SpikingJelly library for SNNs \cite{doi:10.1126/sciadv.adi1480}. SANSAC in alignment with SAC does not utilize any additional encoding or decoding methods.

Artificial neural network agents in RL are fed state variables as described by MDPs where the network is typically trained through a combination of gradient descent and activation functions. However, SNNs are unable to take advantage of gradient descent due to the nonlinear and non-differentiable nature of spike outputs. The discontinous nature leads to the adoption of surrogate gradients \cite{DBLP:journals/corr/abs-1901-09948}. The surrogate gradient has been pivotal to the work on RL in SNNs, allowing for improved performance in complex and continuous environments. The surrogate gradient enables familiar gradient backpropagation seen in traditional neural networks. SANSAC implements the sigmoid surrogate gradient as described by equation \ref{eq:surr}.
\begin{equation}
    \sigma(x, \alpha) = \frac{1}{1+\exp(-\alpha x)}
    \label{eq:surr}
\end{equation}
The surrogate gradient creates network behavior closer to recurrent neural networks by applying backpropagation through an unrolled network with backpropagation through time (BPTT). Aside from the surrogate gradient, all loss function calculations remain the same as SAC.

\section{Experiments}
\label{Experiments}
\subsection{Experimental Settings}
Both the actor and critic networks used in SANSAC and SAC utilized three layers with 2 input hidden dimensions. The same values for a given hidden dimension were used in both each agent's actor and critic networks.
\autoref{tab:Hidden-Dimensions} demonstrates the pairs of hidden dimension variables used in comparison of SANSAC and SAC for quick reference.
\begin{table}[h]

\vskip 0.15in
\begin{center}
\begin{small}
\begin{sc}
\begin{tabular}{lcr}
\toprule
& Hidden Dimension 1 & Hidden Dimension 2 \\
\midrule
\speciallabel{row1}&$256$&$256$ \\
\speciallabel{row2}&$200$&$200$ \\
\speciallabel{row3}&$128$&$128$  \\
\speciallabel{row4}&$64$&$64$ \\
\bottomrule
\end{tabular}

\end{sc}
\end{small}
\end{center}
\caption{Hidden Dimensions}
\label{tab:Hidden-Dimensions}
\vskip -0.1in

\end{table}

All agents were trained for a maximum of 1200 episodes. Training was terminated early if no improvement in episodic reward was observed for 100 consecutive episodes after a minimum of 50 episodes had been completed. This early stopping procedure was applied identically to both SANSAC and SAC to prevent unnecessary computation when learning had stalled. 
\subsubsection{Seed Generation}
Seed generation for all experiments was done randomly to eliminate bias. SAC and SANSAC seeds were kept in sync to each other for reproducibility.

\subsection{Comparison of Algorithms}
\begin{figure}[h]
    \centering
    \includegraphics[width=0.3\linewidth]{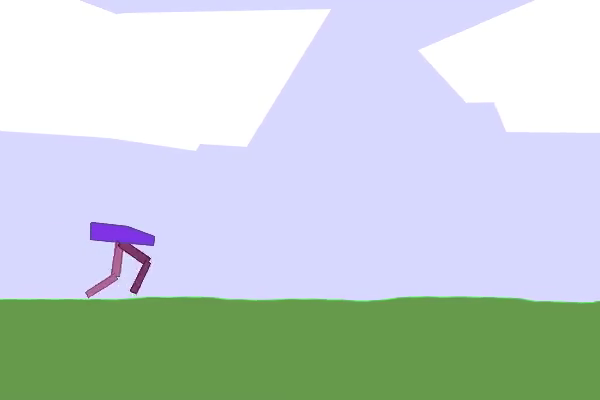}
    \hspace{0.02\linewidth}
    \includegraphics[width=0.3\linewidth]{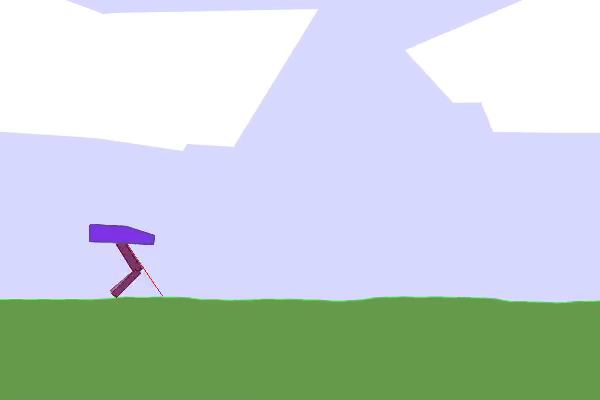}
    \hspace{0.02\linewidth}
    \includegraphics[width=0.3\linewidth]{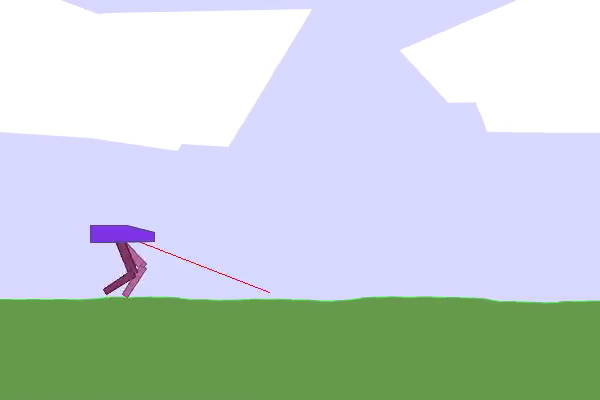}
    \caption{Gymnasium Bipedal Walker: Task is considered solved when 300 points are accumulated in 1600 steps}
    \label{fig:env}
\end{figure}

The evaluation of SANSAC and SAC is done through the Bipedal Walker environment from Farama Foundation's Gymnasium \cite{towers2024gymnasium} as pictured in \autoref{fig:env}. 

We consider the standard baseline performance of both algorithms to be the set of highest hidden dimensions presented in \autoref{tab:Hidden-Dimensions} (\ref{row1}) as it is close to typical hidden dimension settings commonly used and yields the most stable learning out of our tests.

\begin{table}[!h]
\centering
\small
\begin{tabular}{lcccccc}
\toprule
Dimensions & \multicolumn{2}{c}{Mean Reward $\pm$ SD} & \multicolumn{2}{c} {Success Rate} & \multicolumn{2}{c}{Statistical Tests} \\
\cmidrule(lr){2-3} \cmidrule(lr){4-6}
 & SANSAC & SAC &SANSAC&SAC& Mann-Whitney P & Cohen's d \\
\midrule
(\ref{row1}) & $273.9 \pm 130.8$ & $224.5 \pm 162.9$ &70\% & 80\% & 0.7913&0.080\\
(\ref{row2}) & $197.3 \pm 178.2$ & $259.8 \pm 146.2$ &60\% & 80\% & 0.6232& -0.140 \\
(\ref{row3}) & $210.9 \pm 185.9$ & $158.7 \pm 213.5$ &70\% &60\% &0.6776&0.120  \\
(\ref{row4}) & $26.8 \pm 149.4$ & $41.8 \pm 138.5$ &10\% &10\% &0.5205&-0.180  \\
\bottomrule
\end{tabular}
\vspace{0.2cm}
\caption{Performance comparison between SAC and SANSAC. Success Rate indicates percentage of agents achieving reward $\ge 300$ in tests. Statistical tests show no significant differences between algorithms across all dimensions (Mann-Whitney p $>0.05$). Cohen's d effect sizes: $|d|<0.2$ indicates negligible difference, $0.2 \leq |d| < 0.5$ small, $0.5 \leq |d|<0.8$ medium. All configurations show negligible to small effect sizes confirming the practical equivalence.}
\label{tab:performance} 

\end{table}

\begin{figure}[h]
    \centering
    \includegraphics[width=0.8\linewidth]{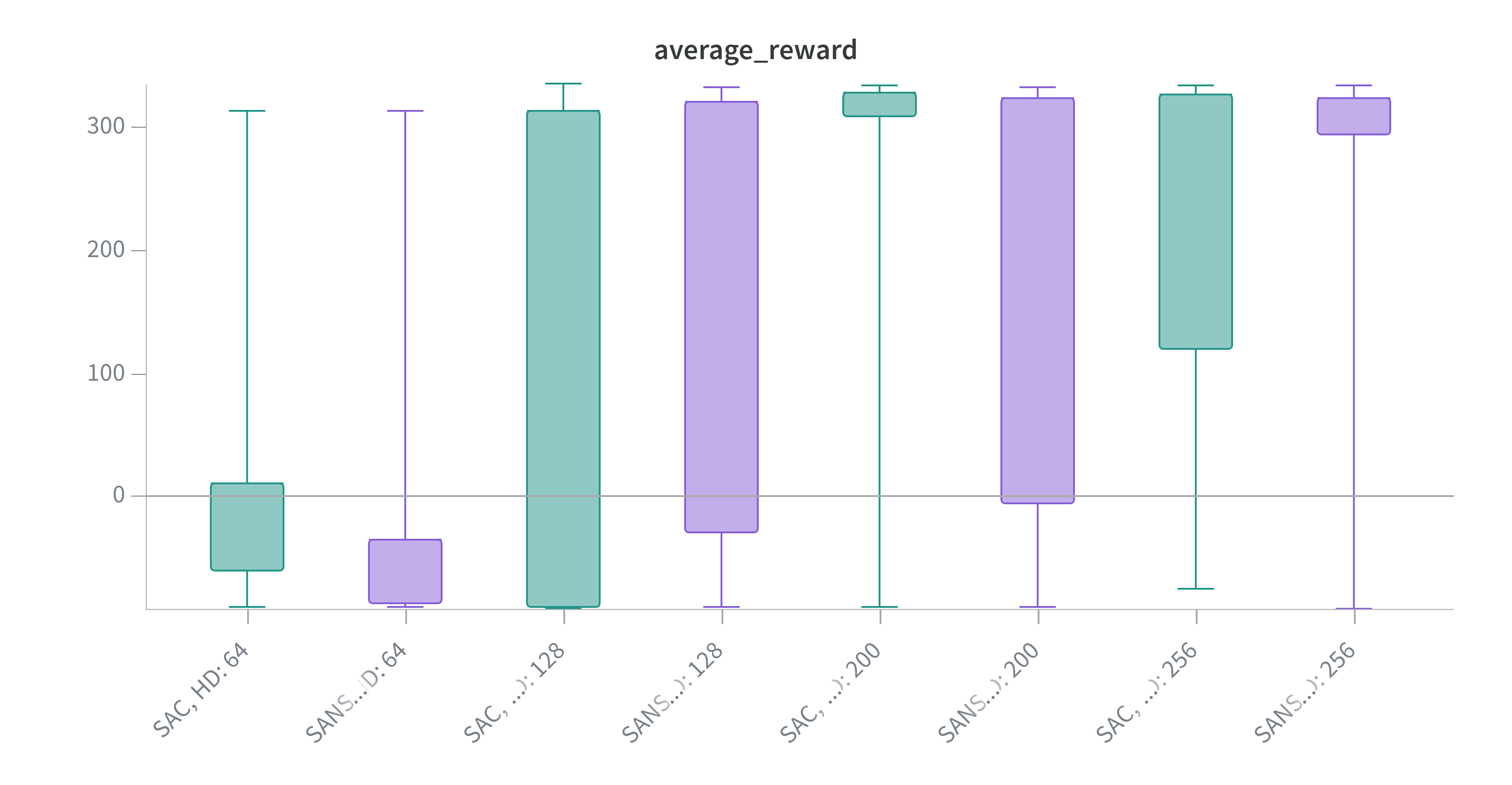}
    \hspace{0.02\linewidth}
    \caption{Box plots demonstrating the average reward of agents in testing. Configurations are labeled to separate SANSAC from SAC in each dimension test.}
    \label{fig:boxplots}
\end{figure}

Comparing the performance of SANSAC and SAC in all hidden dimensions, we observe similar results for both algorithms. Through the use of Mann-Whitney U tests as shown in \autoref{tab:performance} we observe no statistically significant differences between the two algorithms in any configuration.

\begin{figure*}[ht]
  \centering
  \begin{minipage}[t]{0.48\textwidth}
    \centering
    \includegraphics[width=\linewidth,trim={0 0 3mm 5mm},clip]{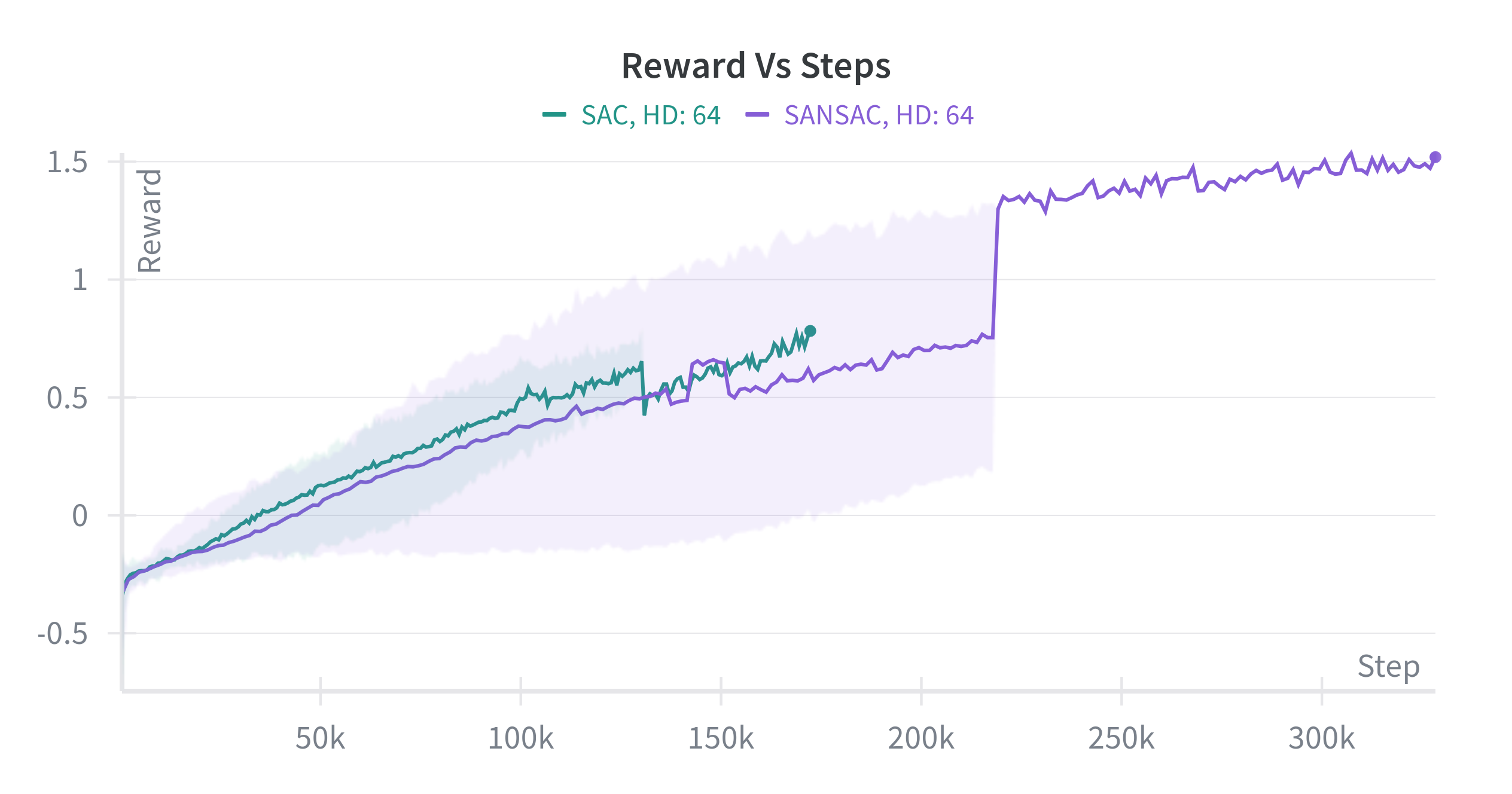}
    \subcaption{Hidden dimension 64}
    \label{fig:64reward}
  \end{minipage}
  \hfill
  \begin{minipage}[t]{0.48\textwidth}
    \centering
    \includegraphics[width=\linewidth,trim={0 0 3mm 5mm},clip]{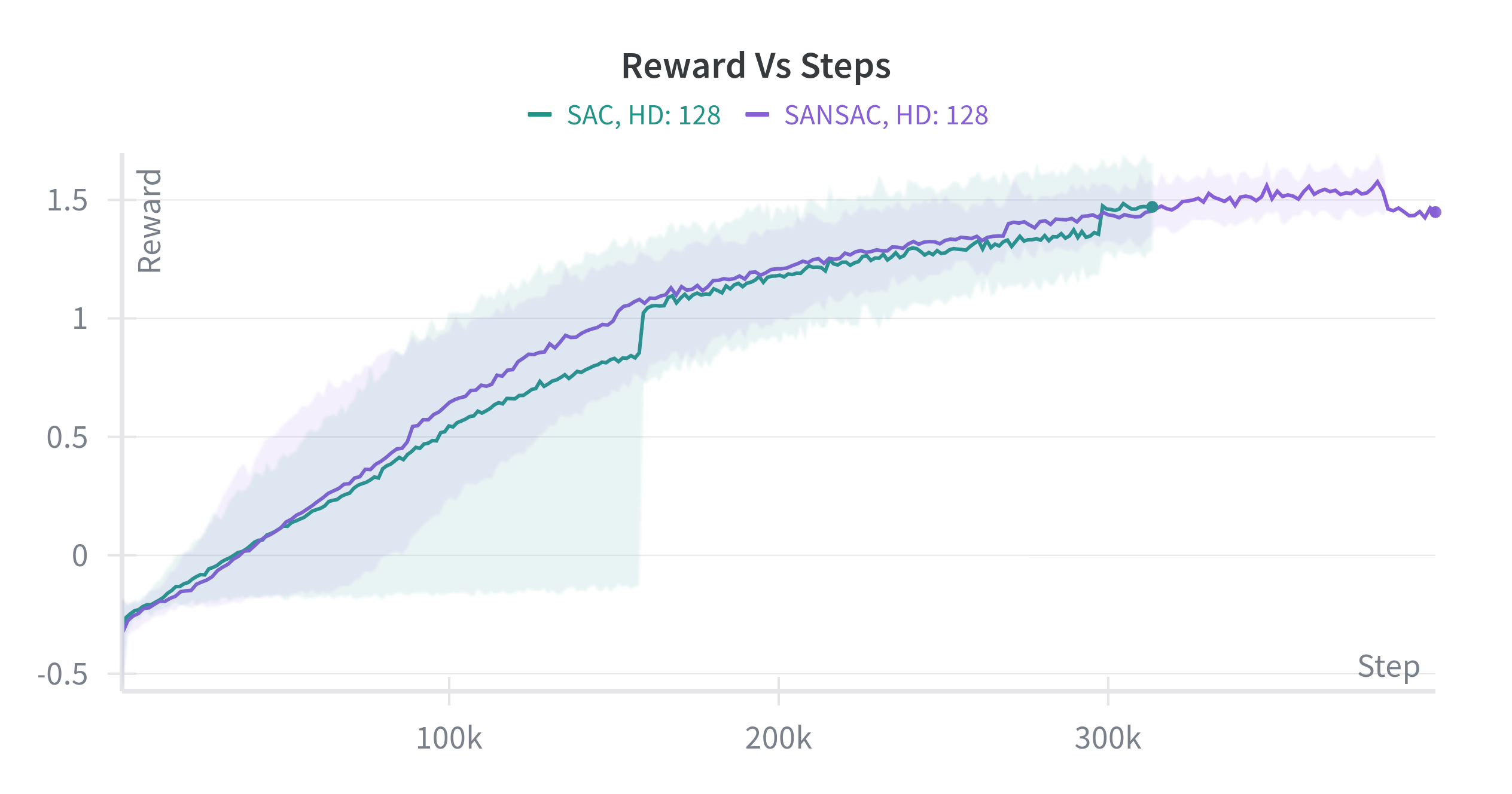}
    \subcaption{Hidden dimension 128}
    \label{fig:128reward}
  \end{minipage}
  
  \vspace{0.5cm} 
  
  \begin{minipage}[t]{0.48\textwidth}
    \centering
    \includegraphics[width=\linewidth,trim={0 0 3mm 5mm},clip]{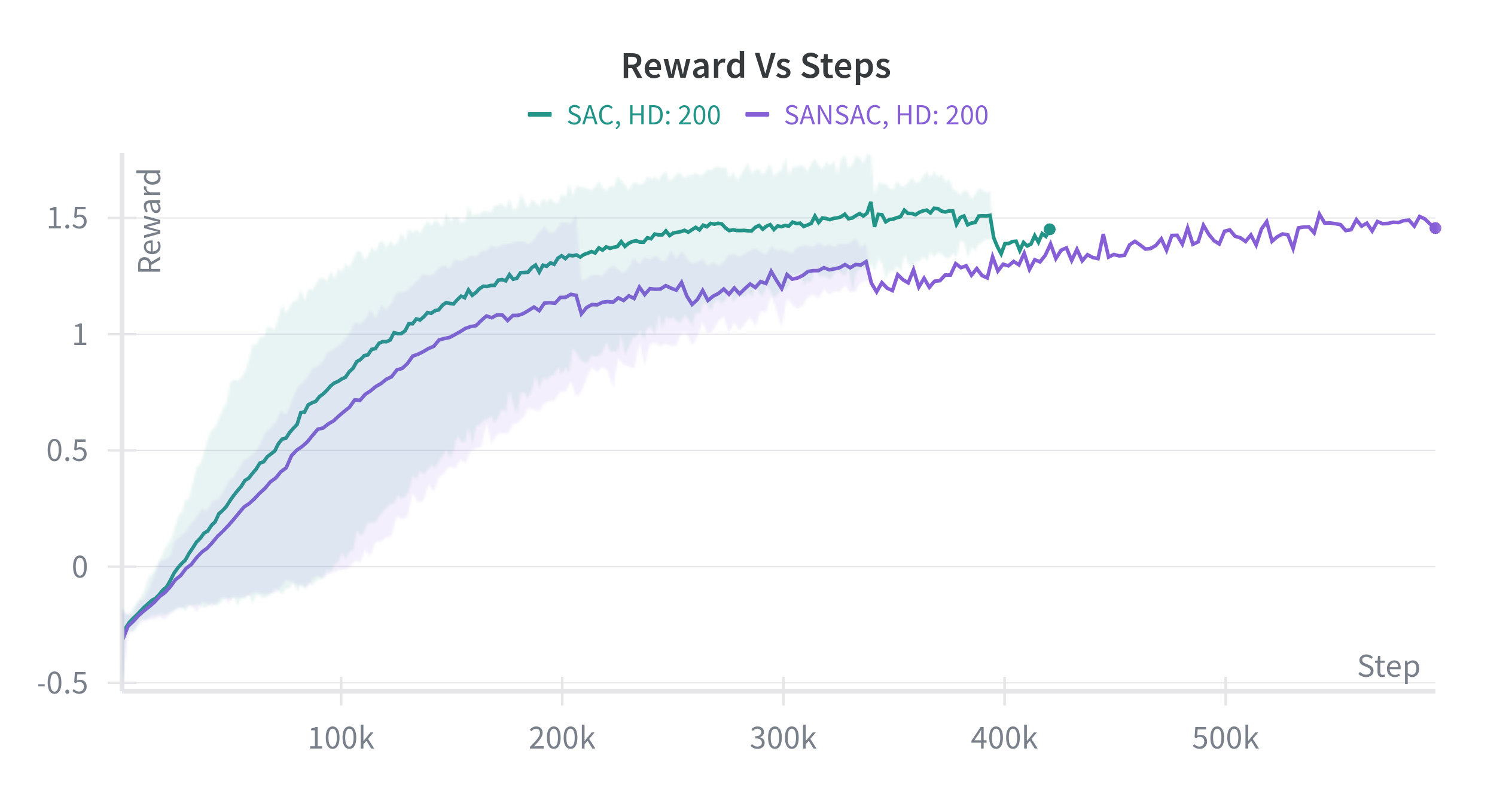}
    \subcaption{Hidden dimension 200}
    \label{fig:200reward}
  \end{minipage}
  \hfill
  \begin{minipage}[t]{0.48\textwidth}
    \centering
    \includegraphics[width=\linewidth,trim={0 0 3mm 5mm},clip]{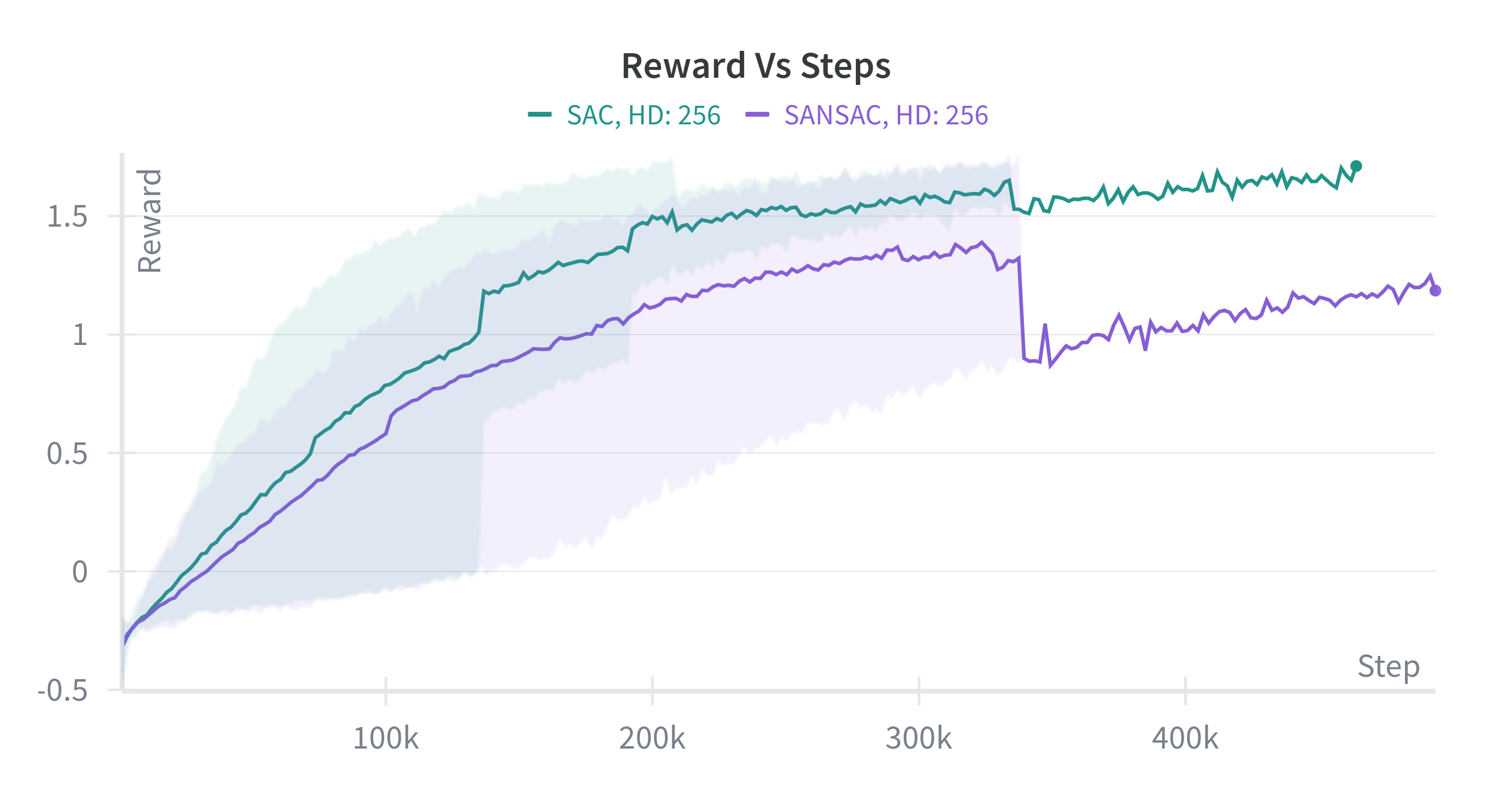}
    \subcaption{Hidden dimension 256}
    \label{fig:256reward}
  \end{minipage}
  
  \caption{Reward progression for SANSAC and SAC across different hidden dimensions. Curves show the mean episodic reward across 10 random seeds. Shaded regions represent the $\pm 1$ standard deviation across seeds. Training was terminated after 1200 episodes or no reward improvement in the last 100 consecutive episodes. Truncated curves indicate early termination either through early convergence or early stopping, not a difference in evaluation protocol.}
  \label{fig:reward}
\end{figure*}

\begin{figure*}[ht]
  \centering
  \begin{minipage}[t]{0.48\textwidth}
    \centering
    \includegraphics[width=\linewidth,trim={0 0 3mm 5mm},clip]{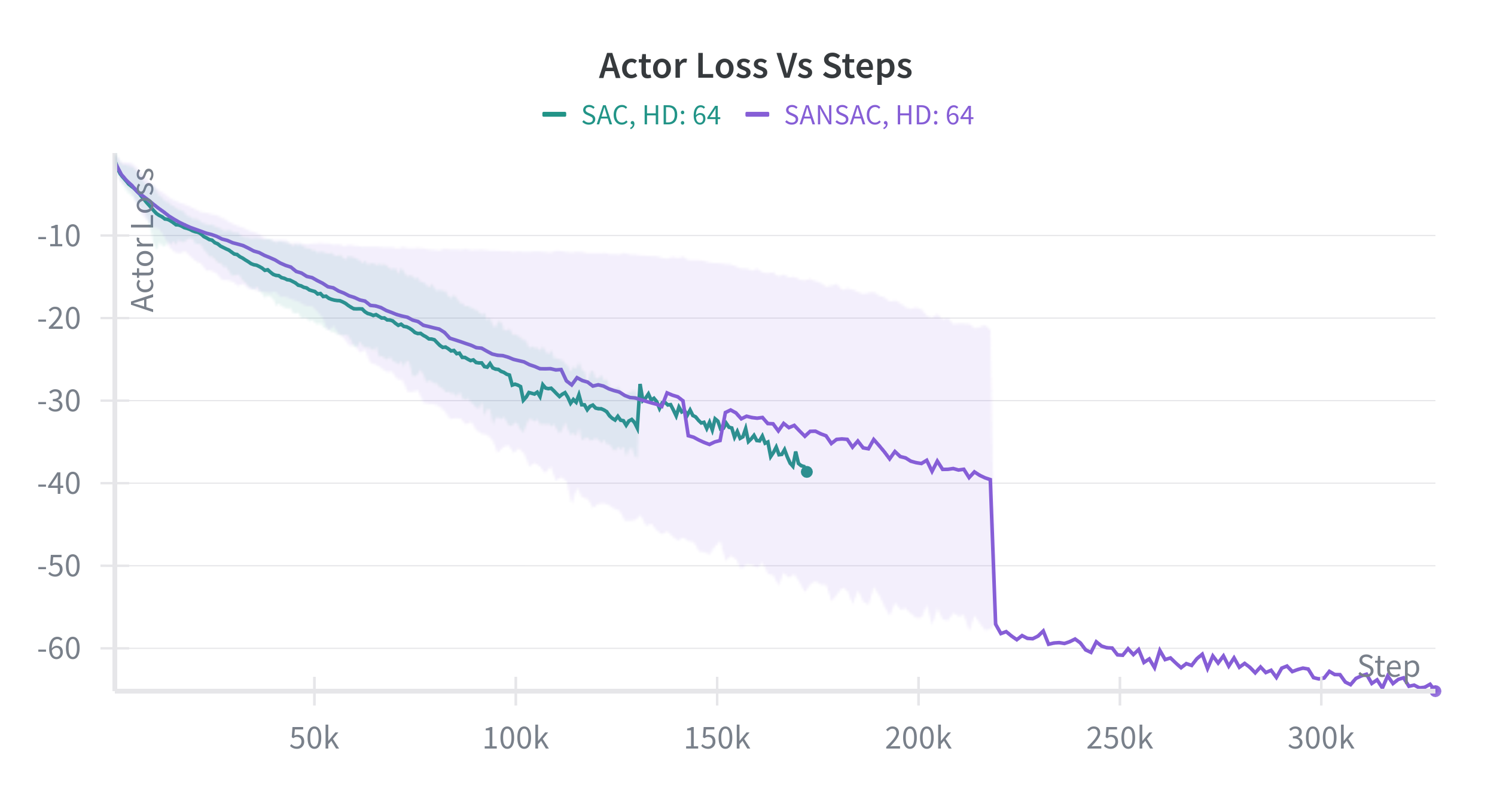}
    \subcaption{Hidden dimension 64}
    \label{fig:64ALoss}
  \end{minipage}
  \hfill
  \begin{minipage}[t]{0.48\textwidth}
    \centering
    \includegraphics[width=\linewidth,trim={0 0 3mm 5mm},clip]{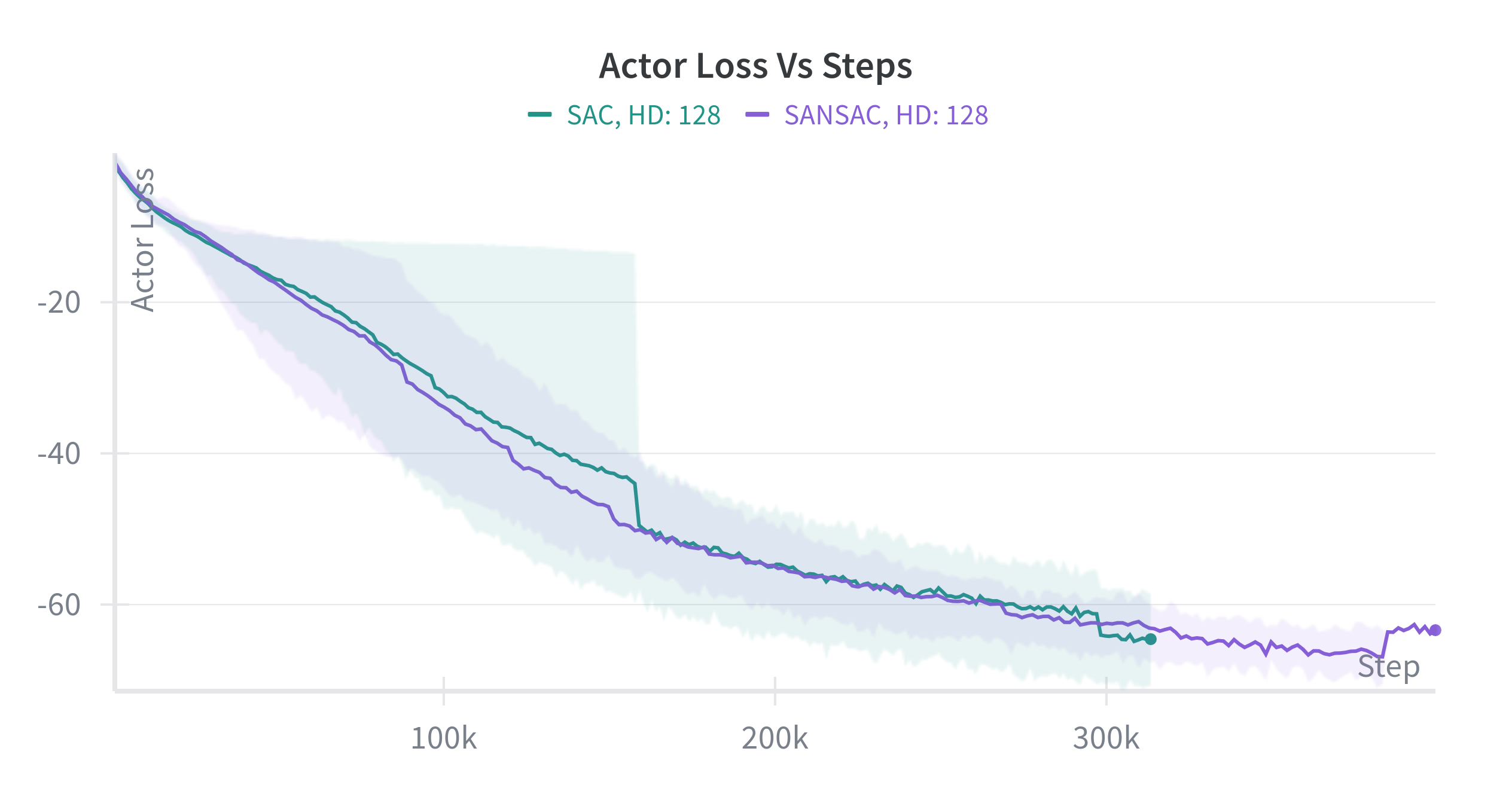}
    \subcaption{Hidden dimension 128}
    \label{fig:128ALoss}
  \end{minipage}
  
  \vspace{0.5cm} 
  
  \begin{minipage}[t]{0.48\textwidth}
    \centering
    \includegraphics[width=\linewidth,trim={0 0 3mm 5mm},clip]{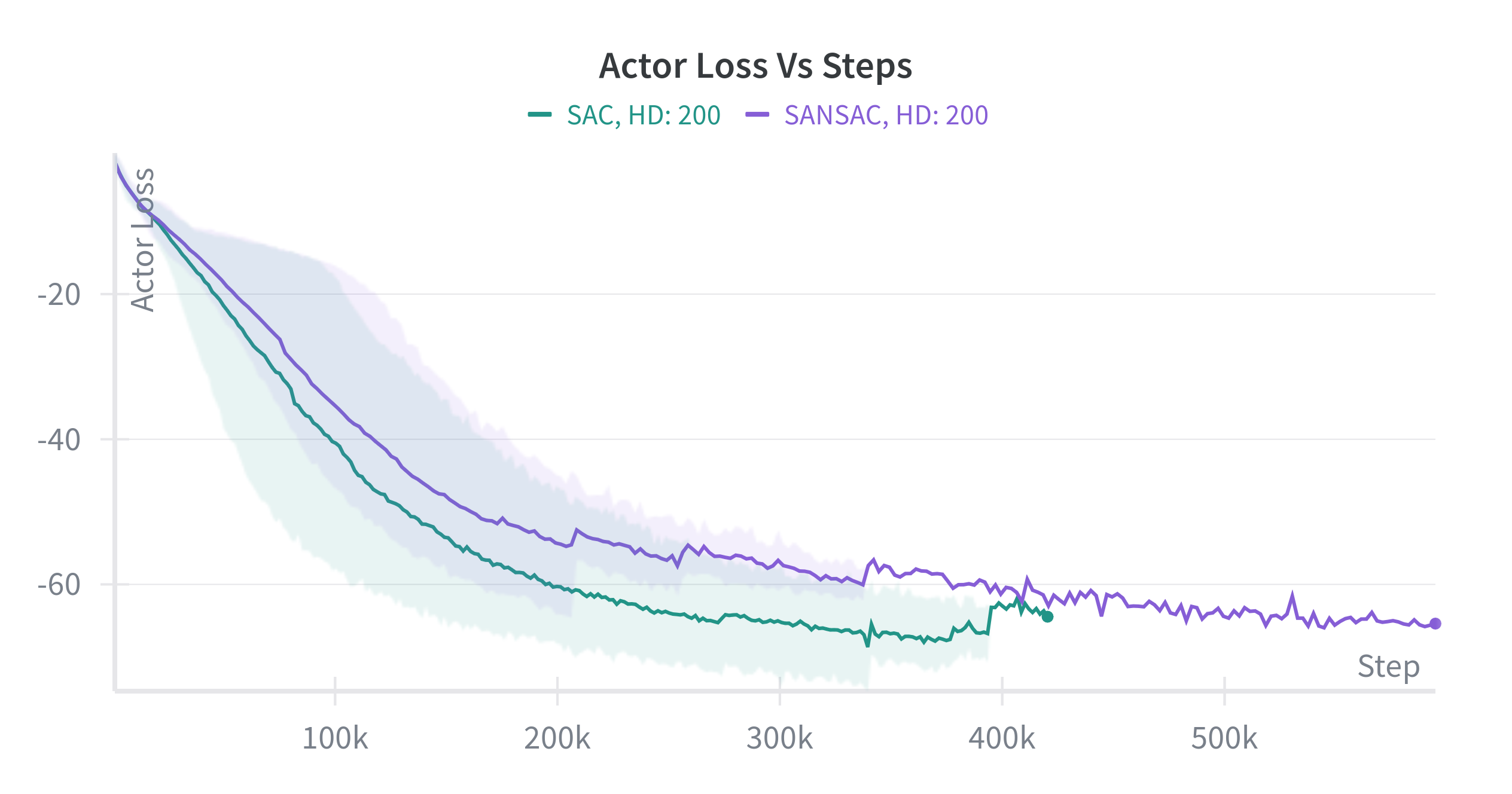}
    \subcaption{Hidden dimension 200}
    \label{fig:200ALoss}
  \end{minipage}
  \hfill
  \begin{minipage}[t]{0.48\textwidth}
    \centering
    \includegraphics[width=\linewidth,trim={0 0 3mm 5mm},clip]{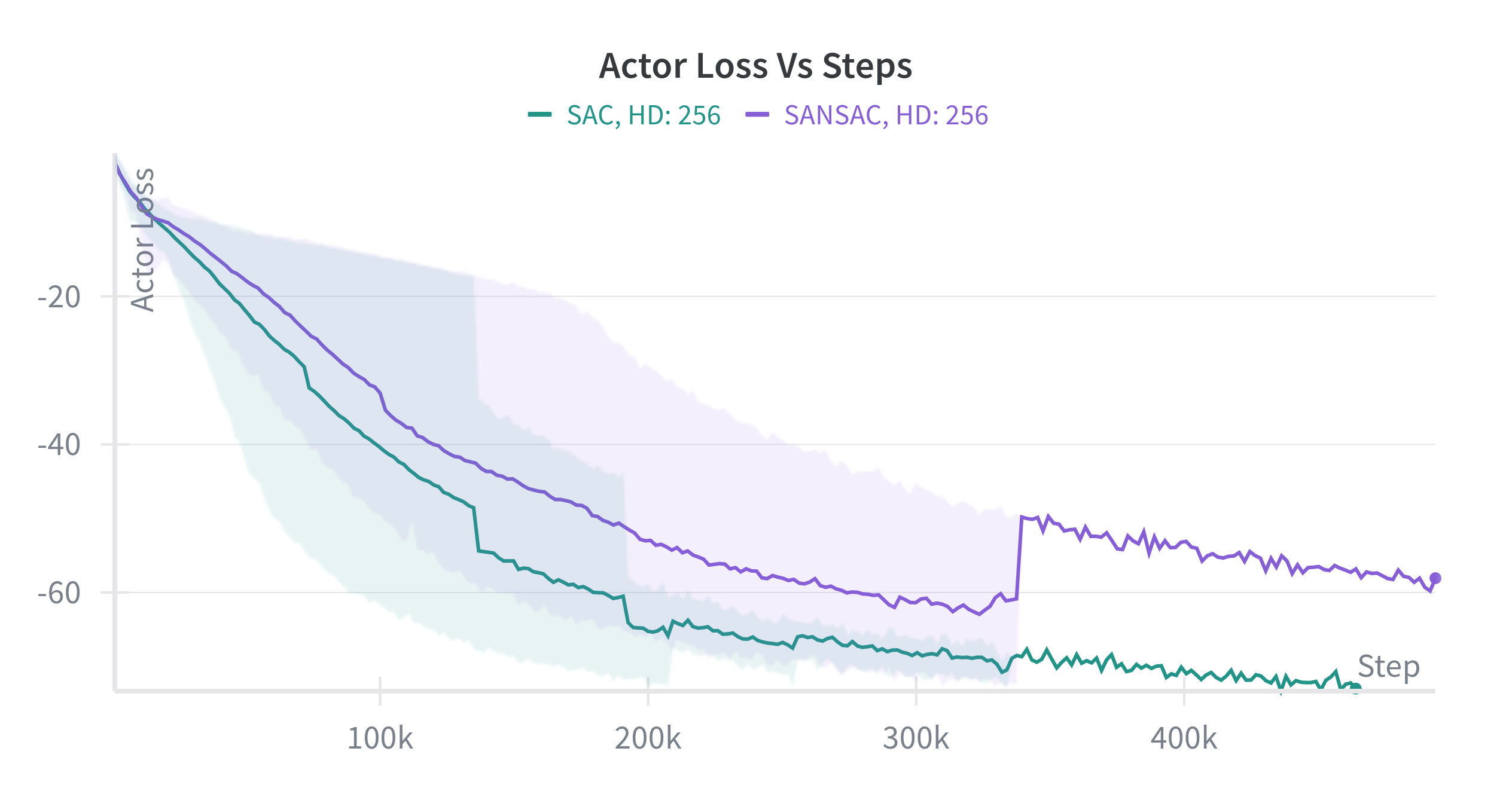}
    \subcaption{Hidden dimension 256}
    \label{fig:256ALoss}
  \end{minipage}
  
  \caption{Actor Loss across steps for agent training averaged across 10 runs of random seeds. Curves show the mean episodic loss across 10 random seeds. Shaded regions indicate $\pm 1$ standard deviation across seeds. The inverse relationship between loss and reward is expected as lower actor loss corresponds to higher expected returns.}
  \label{fig:actorloss}
\end{figure*}

We demonstrate the reward curves across several hidden dimensions in \autoref{fig:reward} and the actor loss curves similarly in \autoref{fig:actorloss}, revealing similar learning dynamics between algorithms. We include both for completeness to provide a complete picture of learning dynamics. The graphs reveal consistent patterns: both algorithms struggle at the lowest dimension (\ref{row4}), improved performance at intermediate dimensions (\ref{row2}, \ref{row3}), and standard stable performance at higher dimensions (\ref{row1}). The similarity present in the convergence and shapes of the reward curves indicate that SANSAC learns similar behavior to that of SAC. Despite this pattern SANSAC's behavior observed in \autoref{fig:256reward} show an instability as reward drops compared to SAC.

Performance degradation as overall topology of the network decreases is an expected result as the decrease in neurons directly leads to a lowered capacity to learn. The lowest dimension (\ref{row4}), both algorithms exhibit this performance drop with only 10\% success rates. As the neurons present increases through higher hidden dimensions, both algorithms show substantial improvement with SANSAC showing performance competitive to SAC.

Through statistical analysis it is observed that SANSAC has an overall higher variance in performance compared to SAC in (\ref{row2}) and (\ref{row3}). Despite this, it is clear that SANSAC can perform on par with SAC.

Notably within our experiments SANSAC demonstrates a consistently longer time required to train an agent. SANSAC takes ~2x the amount of program time for the same amount of episodes in the tested configurations. This difference in training time is shown in \autoref{tab:time_comparison}. This behavior is expected as spiking neuron dynamics require temporal unrolling, creating a bottleneck on conventional hardware which operates sequentially. On neuromorphic hardware, this unrolling would be parallelized creating potential room for speedup. We do not document energy comparison between SANSAC and SAC in conventional hardware as in our experiments reporting was inconsistent, hard to isolate, and overall inconsistent.

\begin{table}[htbp]
  \centering
  \begin{tabular}{lcc}
    \toprule
    & \multicolumn{2}{c}{\textbf{Training Time (minutes)}} \\
    \cmidrule(lr){2-3}
    \textbf{Hidden Dimension} & \textbf{SANSAC} & \textbf{SAC} \\
    \midrule
    (\ref{row4})& 44 & 10 \\
    (\ref{row3}) & 66 & 26 \\
    (\ref{row2}) & 80 & 23 \\
    (\ref{row1}) & 54 & 24 \\
    \midrule
    \textbf{Mean} & \textbf{61} & \textbf{21} \\
    \bottomrule
  \end{tabular}
  \vspace{0.2cm}
  \caption{Average time taken by each configuration to train the agent across 10 random seeds. }
\label{tab:time_comparison}
\end{table}

\section{Limitations}
Our tests of SANSAC demonstrated competitive performance against SAC in multiple hidden dimension configurations. However, several limitations warrant consideration in interpreting the presented results. The comparison of both algorithms was primarily optimized and tested for the Bipedal Walker environment. While this environment represents a challenging continuous control task, the generalizability of our findings to other continuous environments remains to be validated through future experimentation. Additional findings may be captured through testing different environments which may not be captured in current analysis.

The observed computational time penalty for SANSAC is likely a result of temporal unrolling and bottlenecks inherent to simulating a neuromorphic designed algorithm on conventional architecture. However, we cannot definitively quantify what portion of the observed overhead is fundamental to the algorithm versus the implementation-specific inefficiencies in the current set of experiments. The performance of SANSAC on neuromorphic hardware remains an open question not answered through this simulation-based study.

The observed variance in success rates of SANSAC compared to SAC suggest higher sensitivity, potentially due to the stochastic nature of spiking neurons or the surrogate gradient approximation. This sensitivity also affected early stopping behavior as a higher variance could delay convergence detection. This creates additional challenges for achieving consistent results and may affect scalability to larger-scale problems.

\section{Conclusion}
The integration of neuromorphic hardware and its associated design techniques remain underutilized in RL architectures maintained in conventional computing. We introduced SANSAC, a Spiking Actor Network variant of the Soft Actor Critic framework designed to bridge this gap while isolating key behaviors and dynamics necessary to document prior to neuromorphic hardware deployment. Our evaluation demonstrates that SANSAC achieves performance statistically indistinguishable from SAC across multiple hidden dimension configurations. While SANSAC incurs computational overhead during simulation on conventional hardware, this presents a necessary trade-off for maintaining compatibility with neuromorphic hardware architectures where the design would yield energy efficiency benefits.

The success of SANSAC in learning complex continuous control tasks validates the viability of spiking neural networks for reinforcement learning applications. Our findings show that the constraints imposed by neuromorphic design such as surrogate gradients and temporal dynamics do not inherently limit performance compared to traditional deep networks when implemented. These results provide a foundation for future work on neuromorphic reinforcement learning, demonstrating that algorithm design for emerging hardware platforms can maintain competitive performance during the development and training phases conducted on conventional computers. The path toward energy-efficient, neuromorphic RL systems appears viable, with SANSAC serving as a proof of concept for this approach.

\section{Future Work}
Improvement upon this work would be through rigorous testing against other behavior affecting variables including neuron type and encoding and decoding behavior. The current implementation uses LIF neurons and Sigmoid surrogate gradients, but exploration of alternative neuron models and gradient approximations could yield performance improvements.

To expand our research into more practical implementations that help bridge the gap between neuromorphic computing and reinforcement learning we hope to see the design and implementation of similar algorithms on physical neuromorphic hardware. While most neuromorphic chips remain highly experimental and difficult to acquire we suggest the implementation of these algorithms on FPGAs. While complex in implementation these offer a relatively unexplored path that is more accessible and highly customizable.

Further investigation into the scalability of SANSAC and similar neuromorphic designed algorithms to more complex environments and larger network architectures would provide valuable insights into its practical limitations and advantages. Additionally, exploration of hybrid training strategies that leverage both conventional and neuromorphic hardware during different phases of learning could help mitigate the computational overhead observed in our experiments.

\section*{Acknowledgements}
We gratefully acknowledge the support of the NMT Research Opportunities Cooperative Program and the Sophomore Research Program by its sponsors: the Air Force Research Laboratory, Sandia National Laboratories, Alumni Donors, and the Offices of Financial Aid and Academic Affairs of New Mexico Tech.

\bibliography{iclr2026_conference}
\bibliographystyle{iclr2026_conference}

\appendix
\section{Appendix}
\subsection{Hyperparameters}
\renewcommand{\thetable}{\Alph{section}.\arabic{table}}
\setcounter{table}{0}
\begin{table}[h]
\centering
\caption{Hyperparameters}
\vspace{1mm}
\label{tab:newHP}
\begin{tabular}{l l| l }
\toprule
\multicolumn{2}{l|}{Parameter} &  Value\\
\midrule
& learning rate & $3 \cdot 10^{-4}$\\
& discount ($\gamma$) &  0.975\\
& replay buffer size & $10^6$\\
& number of samples per minibatch & 256\\
& target smoothing coefficient ($\tau$)& 0.005\\
& Reward Scale & 5 \\
& Action Dimensions & 4 \\
& State Dimensions & 24 \\
& Alpha & 0.2 \\
& surrogate gradient type & sigmoid \\
& SANSAC timesteps (T) & 16 \\
& optimizer & Adam\\
\bottomrule
\end{tabular}
\end{table}

All data gathered was done on a Nvidia GeForce RTX 4070.
\subsection{Policy Log-Probability Derivation}
\label{app:derivation}

Beginning with the original SAC policy log-probability:
\begin{align*}
    \log \pi(a|s) &= \log \mu(u|s) - \sum_{i=1}^{D} \log(1-\tanh^2(u_i)) \\
    &= \log \mu(u|s) - 2\sum_{i=1}^{D} \log(\operatorname{sech}(u_i)) \quad \text{(using identity } 1- \tanh^2(x) = \operatorname{sech}^2(x)\text{)} \\
    &= \log \mu(u|s) - 2\sum_{i=1}^{D} \left[\log 2 - u_i - \log(1+e^{-2u_i})\right] \\
    &= \log \mu(u|s) - 2\sum_{i=1}^{D} \left[\log 2 - u_i - \operatorname{softplus}(-2u_i)\right]
\end{align*}
The derived transformation replaces the unstable $\tanh$ operations with the alternative $\operatorname{softplus}$ function ($\operatorname{softplus}(x) \triangleq \log(1 + e^{x})$), ensuring gradient stability for continuous control tasks during the policy network backpropagation. This avoids vanishing gradients while maintaining differentiability critical for continuous control environments.

\end{document}

%% file: math_commands.tex
\usepackage{amsmath,amsfonts,bm}

\def\eqref#1{equation~\ref{#1}}

\def\1{\bm{1}}

\DeclareMathAlphabet{\mathsfit}{\encodingdefault}{\sfdefault}{m}{sl}
\SetMathAlphabet{\mathsfit}{bold}{\encodingdefault}{\sfdefault}{bx}{n}

